\pdfoutput=1

\documentclass[11pt]{article}

\usepackage{acl}

\usepackage{times}
\usepackage{latexsym}

\usepackage[T1]{fontenc}

\usepackage[utf8]{inputenc}

\usepackage{microtype}

\usepackage{inconsolata}

\usepackage{twemojis}
\usepackage{graphicx}
\usepackage{multirow}
\usepackage{ctable}
\usepackage{arydshln}
\usepackage{algorithm}
\usepackage{algpseudocode}
\usepackage{arydshln}
\usepackage{amsmath}
\usepackage{amsfonts}
\usepackage{csquotes}

\newcommand{\mn}{\textcolor{black}}
\newcommand{\mg}{\textcolor{black}}
\newcommand{\lb}{\textcolor{black}}
\newcommand{\mc}{\textcolor{black}}

\newcommand{\yes}{\scalebox{1.25}{\twemoji{check mark}}}
\newcommand{\no}{\scalebox{1.25}{\twemoji{cross mark}}}
\newcommand{\taskone}{\scalebox{1.3}{\twemoji{keycap: 1}}}
\newcommand{\tasktwo}{\scalebox{1.3}{\twemoji{keycap: 2}}}
\newcommand{\taskthree}{\scalebox{1.3}{\twemoji{keycap: 3}}}

\usepackage{lipsum}

\title{\textsc{SBAAM}! Eliminating Transcript Dependency \\ in Automatic Subtitling}

\author{Marco Gaido, Sara Papi, Matteo Negri, Mauro Cettolo, Luisa Bentivogli \\
  Fondazione Bruno Kessler \\
  Trento, Italy \\
  \texttt{\{mgaido,spapi,negri,cettolo,bentivo\}fbk.eu} \\}

\begin{document}
\maketitle
\begin{abstract}

Subtitling plays a crucial role in enhancing the accessibility of audiovisual content and encompasses three primary subtasks: translating spoken dialogue, segmenting translations into concise textual units, and estimating timestamps that govern their on-screen duration. Past attempts to automate this process rely, to varying degrees, on automatic transcripts, employed diversely for the three subtasks. In response to the acknowledged limitations associated with this reliance on transcripts, recent research has shifted towards transcription-free solutions for translation and segmentation, leaving the direct generation of timestamps as uncharted territory. To fill this gap, we introduce the first direct model capable of producing automatic subtitles, entirely eliminating any dependence on intermediate transcripts also for timestamp prediction. Experimental results, backed by manual evaluation, showcase our solution's new state-of-the-art performance across multiple language pairs and diverse conditions.
\end{abstract}

\section{Introduction}

\begin{figure*}
    \centering
    \includegraphics[width=\linewidth]{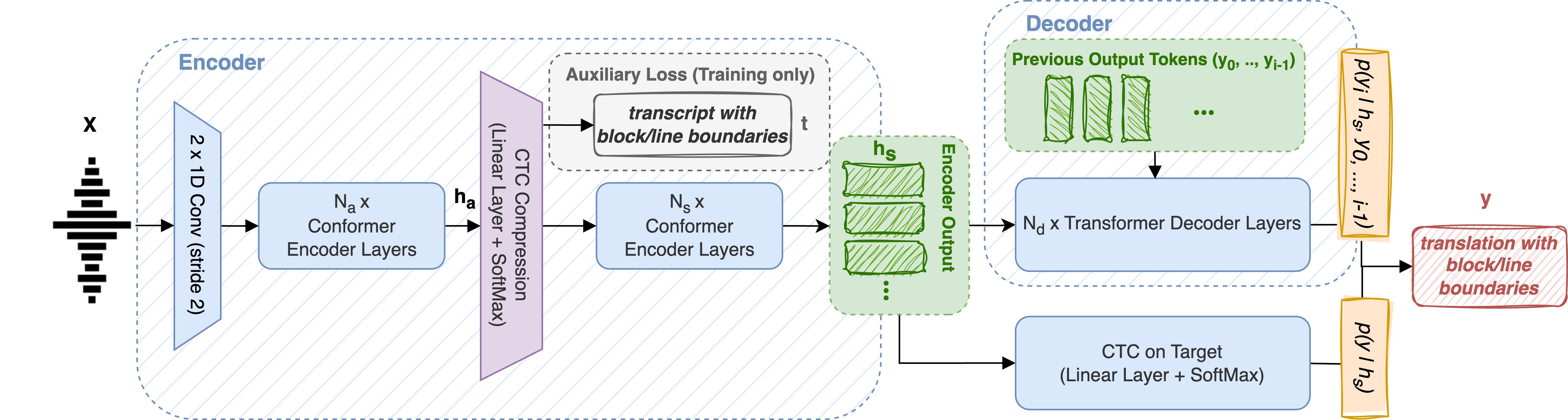}
    \caption{Architecture of our model.}
    \label{fig:arch}
\end{figure*}


Subtitling aims to facilitate the accessibility
of audiovisual media, such as movies, TV shows, and video lectures, by providing users with a textual translation of spoken content.
Subtitles consist of two components: a textual block, typically encompassing one or two lines, and its corresponding time duration, indicated by start and end timestamps. To ensure effective on-screen presentation and minimize users' cognitive load, subtitles should conform to spatio-temporal constraints \citep{bogucki2004constraint,khalaf2016introduction}. These include restrictions on the maximum number of characters per line and a display duration that guarantees synchronization with the video while granting
viewers sufficient time to read the entire text.

Automating the task involves addressing three main subtasks: \taskone{} translation of the spoken content, \tasktwo{} segmentation of the translated text into blocks and lines,
and \taskthree{}~estimation of the timestamps for each block.
Early approaches to automatic subtitling (AS) adopted a cascade architecture \citep{piperidis-etal-2004-multimodal,oliver-gonzalez-2006-automatic,10.1016/j.specom.2017.01.010,bojar-etal-2021-elitr}, i.e. a pipeline of components, including automatic speech recognition (ASR) and machine translation (MT) models. In these systems, transcripts served as the foundational element for all three subtasks, despite the well-documented limitations of this reliance on them,
such as error propagation \citep{sperber-paulik-2020-speech}, the loss of useful prosody information \citep{isometric2022,tam22_interspeech}, inapplicability to source languages lacking written forms \citep{lee-etal-2022-textless}, and higher computational and environmental cost \citep{strubell-etal-2019-energy} due to the need to run multiple models.


%
%
To cope with these limitations, subsequent studies aimed to streamline the subtitling pipeline 
by 
reducing its dependency on transcripts.
%
%
In this endeavor, direct speech-to-text translation (ST) systems \citep{berard_2016, weiss2017sequence}, capable of translating speech without recourse to intermediate symbolic representations, were successfully used by \citet{karakanta-etal-2020-42} for the translation step (subtask~\taskone).
%
%
%
For subtitle segmentation (subtask~\tasktwo), efforts focused on adapting MT \citep{etchegoyhen-etal-2014-machine,smtsubtitling,volk-etal-2010-machine,matusov-etal-2019-customizing,koponen-etal-2020-mt,cherry21_interspeech} and language models \citep{ponce-etal-2023-unsupervised} to directly produce translations 
\lb{that incorporate}
block and line
boundaries as specific tags, also by exploiting audio information \citep{papi-etal-2022-dodging}.

To date, compared to subtasks \taskone{} and \tasktwo, timestamp estimation (subtask~\taskthree) has received much less attention. This 
regards
not only 
the elimination of intermediate transcription steps, the primary focus of this work, but also 
evaluation, 
as there is still no reliable and informative metric for directly assessing the quality of the predicted timestamps.
%
%
As for the task itself, current approaches, including those exclusively based on direct ST models \citep{10.1162/tacl_a_00607,papi-etal-2023-direct,bahar-etal-2023-speech}, still require transcripts for timestamp estimation, which involves \textit{i)} generating captions (i.e., segmented transcripts), \textit{ii)} estimating timestamps for caption blocks, typically through a Connectionist Temporal Classification (CTC) loss \citep{ctc-2006}, and \textit{iii)} projecting them onto target subtitles. 
%
%
Regarding automatic evaluation, current metrics \citep{wilken-etal-2022-suber} are by design holistic and therefore inadequate to precisely measure timestamp estimation quality.
%
%
%
%
%

To bridge these gaps, we introduce and 
evaluate
the first fully end-to-end AS solution\footnote{\label{fn:open}All our code and pre-trained models are available at \url{https://github.com/hlt-mt/FBK-fairseq/} under Apache Licence 2.0.} that seamlessly produces both subtitles (i.e., segmented translations) and their timestamps 
without any reliance on intermediate transcripts.
This innovation is realized by incorporating into our model the capability to directly determine the temporal alignment between 
the spoken utterances and their corresponding translation in subtitle form.
Along this direction,  
our contributions are the following:
\smallskip
   
    \noindent $\bullet$ \textbf{We propose two methods for timestamp estimation} (\S\ref{sec:timestamp-est}), respectively based on applying the CTC \lb{loss}
    directly on translations
    \citep{pmlr-v162-zhang22i}, and on estimating the audio-text temporal alignment from the attention mechanism \cite{papi-etal-2023-attention,alastruey2024speechalign}. Both approaches are complemented by the 
    joint CTC decoding \citep{8068205}, which 
    yields higher translation quality and a more precise alignment between the generated text and the corresponding audio \citep{yan-etal-2023-ctc};
    \smallskip
    
    \noindent $\bullet$ \textbf{We introduce SubSONAR,\footnote{Available at \url{https://github.com/hlt-mt/subsonar/} under Apache License 2.0 and on PyPi (\url{https://pypi.org/project/SubSONAR/}).} a novel metric for evaluating timestamp quality} (\S\ref{sec:subsonar}),
    which is based on SONAR \citep{Duquenne:2023:sonar_arxiv} and designed to be sensitive to time shifts, enabling a focused evaluation of 
   timestamps;
    \smallskip

\noindent $\bullet$ \textbf{We validate our approach through 
comparative
experiments} (\S\ref{sec:results}) on 7 language pairs, 
2
data conditions, and 4 domains, achieving new state-of-the-art results on different benchmarks and demonstrating better performance compared to cascade architectures for automatic subtitling;
\smallskip

    \noindent $\bullet$  \textbf{We verify the efficacy of our model through 
    manual evaluation} (\S\ref{sec:manual-eval}),
    attesting a significant reduction in timestamp adjustments of $\sim$24\% compared to the previous state of the art.

\section{Direct Model for Automatic Subtitling}
\label{sec:model}

Following previous work on direct ST \citep{liu2020bridging,xu-etal-2021-stacked,DBLP:conf/ijcai/0008YDZKWXZ23,wu2023decoderonly}, we build a direct autoregressive encoder-decoder model, where the encoder is composed of three blocks: \textit{i)} an acoustic encoder -- made of two 1D convolutions with stride 2 and $N_a$ Conformer \citep{gulati20_interspeech} layers, \textit{ii)} a length adaptor -- a CTC Compression \citep{gaido-etal-2021-ctc} module that averages the vectors corresponding to the same CTC prediction, and \textit{iii)} a semantic encoder -- made of $N_s$ Conformer layers.
The encoder output is then fed to an autoregressive decoder $\mathcal{D}$ and, in parallel, to a CTC on Target (TgtCTC) module \citep{pmlr-v162-zhang22i,yan-etal-2023-ctc}.
The full architecture is shown in Figure \ref{fig:arch}.

We train our model with a composite loss ($\mathcal{L}$), which is a linear combination of a label smoothing cross-entropy (CE) loss \cite{szegedy2016rethinking} on the decoder $\mathcal{D}$, a CTC loss on the TgtCTC module, and a CTC loss on the CTC Compression module:
\begin{multline*}
    \mathcal{L} = \lambda_1  \, \text{CTC}(h_a, t) + \lambda_2 \,  \text{CTC}(h_s, y) \\+ \lambda_3  \, \text{CE}(\mathcal{D}(h_s, y), y)
\end{multline*}
where $\lambda_{1,2,3}$ are the loss weights, $h_a$ is the acoustic encoder output, $h_s$ is the encoder output, $y$ is the target subtitle, and $t$ is the caption.
It is noteworthy that the caption ($t$), 
\mn{albeit optionally}
used for training, is not 
\mn{strictly required by}
our model. Additionally, \textbf{captions are neither generated nor utilized by our novel timestamp estimation methods} (\S\ref{sec:timestamp-est}), which exclusively rely on the generated subtitles.
\mn{Indeed,} the auxiliary CTC loss on captions and the CTC compression module can be entirely omitted, at the cost of a slight reduction in quality \citep{pmlr-v162-zhang22i}, 
\mn{and}
captions can be replaced with speech units \citep{hubert,zhang-etal-2022-speechut} as the module target. However, we opt to retain the CTC compression module (as well as the related auxiliary CTC loss on captions), not only for its benefits in terms of computational efficiency and translation quality but, most importantly, to enable the use of the same model with earlier timestamp estimation solutions, which require a CTC module with captions as the target \citep{bahar-etal-2023-speech,10.1162/tacl_a_00607}. This allows a direct comparison between different timestamp estimation methods, employing the same model and generated subtitles.
At inference time, subtitles are generated by estimating their likelihood with the joint CTC/attention framework with CTC rescoring \citep{yan-etal-2023-ctc} through a linear combination of the probabilities of the TgtCTC module and the decoder: 
\begin{equation*}
    p = p_\mathcal{D}(y_i | h_s, y_{0, ..., i-1}) + \alpha \, p_{\text{TgtCTC}}(y_{0, ..., i} | h_s)
\end{equation*}
where $\alpha$ is a hyperparameter. 

\paragraph{Multilinguality.}
We
experiment
\mn{both in}
bilingual 
\mn{(English to a target language)}
and \mn{in} multilingual (English to many languages) settings. In the latter case, we prepend a learned language embedding to the previous output tokens to be fed to the decoder \citep{9003832,wang2020fairseqs2t}. 
In addition, we sum the same learned embedding to all the vectors of the encoder output ($h_s$) before processing them with the TgtCTC module, to inform the module about the language to generate.\footnote{We explored alternative solutions: adding the language embedding at the beginning of the encoder before all Conformer layers, as per \citep{9004003}; adding it to the vectors obtained after the CTC compression module; multiplying the language embedding vector instead of summing it; using separate learned language embeddings for the decoder and the encoder. However, none of them led to better results.}
%
%

\section{Estimation of Block Timestamps}
\label{sec:timestamp-est}

In this section, we present novel methods for estimating the block timestamps without 
\mg{requiring the transcripts.}
These solutions not only avoid error propagation and are applicable to unwritten source languages but also exhibit increased speed compared to current methods, skipping the transcript generation step, thereby minimizing significant overhead. 
\mg{Our}
methods involve either directly applying the CTC-segmentation algorithm \citep{ctcsegmentation} to the CTC on Target module (\S\ref{subsec:subtitle_ctc_segm}) or leveraging the encoder-decoder attention scores to find the audio-text alignment (\S\ref{subsec:att-timestamp}). 

\subsection{Subtitle CTC-based Estimation}
\label{subsec:subtitle_ctc_segm}

The error propagation and approximations caused by the cross-lingual Levenshtein alignment proposed by \citet{papi-etal-2023-direct} can be avoided by directly estimating the timestamps on the target, i.e., on the generated subtitles. This is realized by exploiting the predictions of the TgtCTC module placed on top of the encoder, which are used by the CTC segmentation algorithm \citep{ctcsegmentation}, together with the input audio, to retrieve the timestamp information. 
This subtitle CTC-based method (SubCTC) enables the direct alignment between the input audio features (representing the time over the speech sequence) and the boundaries of the generated subtitle blocks, eliminating the need for intermediate transcript alignments with the audio and their projection into the target side.

The SubCTC method builds upon the assumption that the CTC module can learn meaningful alignments between source audio and target texts, as it does when predicting transcripts \citep{7178778,sainath20_interspeech,chen23c_interspeech} without direct supervision. However, the validity of this assumption has not been verified yet. Rather, related works in non-autoregressive translation showed the inability of such models to implicitly learn complex word reordering \citep{chuang-etal-2021-investigating,Ran_Lin_Li_Zhou_2021,xu-etal-2021-distilled}.
Motivated by this potential issue, we 
\lb{devise}
an alternative method that leverages the cross-attention matrix to infer source-target alignments.

\subsection{Attention-based Estimation}
\label{subsec:att-timestamp}
\lb{Building on}
the extensive literature in text-to-text translation and language modeling discussing the quality of source-target alignments learned by the attention mechanism \citep{tang-etal-2018-analysis,zenkel2019adding,garg-etal-2019-jointly,chen-etal-2020-accurate} and recent works on speech-to-text \citep{papi-etal-2023-attention,alastruey2024speechalign}, we propose to exploit the encoder-decoder attention scores to determine subtitle block timings and devise two different modalities of leveraging this information.
For the first one, we adapt the Dynamic Time Warping (DTW) algorithm \citep{1163055}, which is commonly used to determine token-level timestamps in ASR (\S\ref{subsec:att-timestamp-dtw}).\footnote{For example, the DTW is used in the Transformers implementation of Whisper \citep{whisper}: \url{https://github.com/huggingface/transformers/blob/v4.34.0/src/transformers/models/whisper/modeling_whisper.py\#L1817}.}
For the second one, recognizing that DTW relies on the assumption of monotonic source-target alignments, which does not always hold in our case, we introduce a new algorithm named \textsc{SBAAM} (\textsc{Speech Block Attention Area Maximization}), specifically designed for subtitles (\S\ref{subsec:att-timestamp-baam}).

\subsubsection{Attention-based DTW for Subtitling}
\label{subsec:att-timestamp-dtw}

\begin{algorithm}[t]
\small
\caption{Attention-based DTW}\label{alg:dtw}
\begin{algorithmic}[1]
\State $N, L \gets nTokens, audioLen$
\State $B \gets blockIdxs$ \Comment{List of block boundary indices}
\State $A \gets attnMatrix$ \Comment{$A \in \mathbb{R}^{N \times L}$}
\State $A \gets $\textbf{StdNorm}($A, axis=0$) \Comment{Column-wise normalize}
\State $A \gets $\textbf{MedianFilter}($A, width=7, axis=1$)
\State $D \gets $\textbf{DTWDistance}($-A$) \Comment{Negated attentions as costs}
\State $n, l, P, blockTimings \gets N, L, [(N, L)], [\,]$
\While{$P[-1] \ne (0, 0)$} \Comment{Backtrack best DTW path}
  \If{$n \in B$} 
        \State  $P \gets$ \textbf{Append}$(P,  (n-1, l-1))$ 
        \State  $blockTimings \gets$ \textbf{Append}($blockTimings, l$) 
  \Else 
        \State  $P \gets$ \textbf{Append}$(P, \underset{i \in \{n, n-1\}, j \in \{l, l-1\}}{\mathrm{argmin}} D[i, j])$
\EndIf
\State $n, l \gets P[-1]$ 
\EndWhile \\
\Return \textbf{reverse}($blockTimings$)
\end{algorithmic}
\end{algorithm}

The DTW algorithm is a dynamic programming algorithm similar to Viterbi \citep{6770768}, which finds the best path (i.e., the one that minimizes a distance function) between two temporal sequences with varying speeds. This algorithm operates under the assumption that all the elements of the two sequences are aligned and that the mapping between the two sequences is monotonic. 
In our case, while the assumption may not be applicable at the token (or word) level, it remains valid at the block level, where the order has to be 
maintained; we therefore investigate its applicability to our task.
  
We use the additive inverse of the attention matrix as a distance function and follow the improved DTW algorithm by \citet{6089967} that applies a 1D median filter \citep{1163188}\footnote{A filter that takes the median value within a pre-defined window defined as the width of the filter itself.} over the speech sequence for each text token, after a standard normalization over the text tokens (see Alg. \ref{alg:dtw}). 
After computing the accumulated DTW distance matrix, we define the best path in the backtracking phase, with three possible moves: \textit{i)} moving to the preceding element of the first sequence, \textit{ii)} moving to the preceding element of the second sequence, or \textit{iii)} moving to the preceding elements of both sequences.
In this phase, we force moving across both (speech and text) sequences when the block boundary token (\texttt{<eob>}) is encountered.
This decision is based on two observations: \textit{i)} a manual inspection of attention matrices revealed that the attention of \texttt{<eob>} tokens is often unreliable (e.g., focused on the last token of the speech sequence or silence), and \textit{ii)} we want to ensure that at least one speech segment (equivalent to 40ms) is assigned to each block by skipping the \texttt{<eob>} token in the DTW search (see lines 9-12 of Alg. \ref{alg:dtw}). As a result, we use the timings assigned to the \texttt{<eob>} tokens as the temporal boundaries (timestamp) for the corresponding subtitle blocks.

\subsubsection{\textsc{SBAAM}}
\label{subsec:att-timestamp-baam}

\begin{algorithm}[t]
\small
\caption{SBAAM}\label{alg:baam}
\begin{algorithmic}[1]
\State $N, L \gets nTokens, audioLen$
\State $B \gets blockIdxs$ \Comment{List of block boundary indices}
\State $A \gets attnMatrix$ \Comment{$A \in \mathbb{R}^{N \times L}$}
\State $A \gets $\textbf{StdNorm}($A, axis=0$) \Comment{Row-wise normalize}
\State $A[A < 0] \gets -\epsilon$ \Comment{$\epsilon$ set to $0.01$}
\State $n, l, blockTimings \gets 0, 0, [\,]$
\ForAll{$i_b \in [0, |B|)$} 
  \State $\begin{aligned}l \gets \hspace{-5pt} \underset{j \in (l, L-|B|+i_b)}{\mathrm{argmax}} (& \sum A[n:B[i_b], l:j] + \\[-5pt] & \sum A[B[i_b]+1:N, j+1:L]) \end{aligned}$
  \State $blockTimings \gets$ \textbf{Append}($blockTimings, l$)
  \State $n \gets B[i_b]$
\EndFor \\
\Return $blockTimings$
\end{algorithmic}
\end{algorithm}

\begin{table*}
\small
\setlength{\tabcolsep}{4pt}
    \centering
    \begin{tabular}{l|ccccccc|c||ccccccc|c}
        \multirow{2}{*}{\textbf{Model}} & \multicolumn{8}{c||}{\textbf{SubER} ($\downarrow$)} & \multicolumn{8}{c}{\textbf{SubSONAR} ($\uparrow$)} \\
        \cline{2-17}
         & de & es & fr & it & nl & pt & ro & AVG & de & es & fr & it & nl & pt & ro & AVG \\
         \hline
        LEV & 60.2 & 47.9 & 53.7 & 52.0 & 49.0 & 46.1 & 49.8 & 51.2 & .677 & .692 & .670 & .689 & .688 & .691 & .688 & .685 \\
        \quad - joint CTC & 61.2 & 49.0 & 54.7 & 52.5 & 49.7 & 47.1 & 50.7 & 52.1 & .667 & .693 & .673 & .690 & .683 & .688 & .687 & .683 \\
        \hdashline
        SubCTC & 59.9 & 47.5 & 53.5 & 51.7 & 48.7 & 45.7 & 49.4 & 50.9 & .718 & .739 & .713 & .724 & .731 & .731 & .719 & .725 \\
        ATTN DTW & 59.9 & \textbf{47.5} & \textbf{53.4} & \textbf{51.6} & \textbf{48.6} & \textbf{45.5} & \textbf{49.2} & \textbf{50.8}  & .745 & .775 & .747 & .761 & .758 & .765 & .754 & .758 \\
        \textsc{SBAAM} & \textbf{59.8} & \textbf{47.5} & \textbf{53.4} & \textbf{51.6} & 48.7 & \textbf{45.5} & 49.3 & \textbf{50.8} & \textbf{.749} & \textbf{.780} & \textbf{.753} & \textbf{.765} & \textbf{.767} & \textbf{.770} & \textbf{.761} & \textbf{.764} \\
    \end{tabular}
    \caption{Results of the timestamp estimation methods in terms of SubER (cased) and subSONAR for all the 7 languages of MuST-Cinema.}
    \label{tab:mustcinema-suber-subsonar}
\end{table*}

To relax the constraint of alignment monotonicity, we devise a new algorithm to estimate the timing of block boundaries. Intuitively, our method (see Alg. \ref{alg:baam}) maximizes the attention scores within a rectangular area encompassing the tokens belonging to a block (i.e. all the tokens between two \texttt{<eob>}) along one axis, and the assigned span across the temporal dimension of the speech sequence along the other axis. For this reason, we named it \textsc{\textbf{S}peech \textbf{B}lock \textbf{A}ttention \textbf{A}rea \textbf{M}aximization} or \textsc{SBAAM}. 
The \textsc{SBAAM} algorithm encompasses two steps.
First, the attention matrix is normalized (following the previous procedure) and all negative values are set to a small negative value ($-\epsilon$). This is required as the attention values are typically peaky in the range $[0, 1]$ due to the \texttt{softmax}, with a few large values and random fluctuations close to 0 for all the others. 
After normalization, some small values may become very negative, while others may be closer to 0, even though both indicate low attention. Despite this, we maintain them as negative values to penalize the integration of areas with very low attention.
Second, for each generated subtitle block (i.e., for each \texttt{<eob>}), we iteratively determine the timing by selecting the splitting point that maximizes the area of the first block with the audio up to that point and the rest of the text with the remaining audio. At the end of this process, we obtain the start and end timing (timestamp) for all blocks.

\section{SubSONAR}
\label{sec:subsonar}
Since it was proposed, the SubER\footnote{In this work, we compute SubER-cased unless specified otherwise as per \citep{papi-etal-2023-direct}.} metric \citep{wilken-etal-2022-suber} has been used to provide a holistic evaluation of subtitles, encompassing 
translation quality, block/line segmentation accuracy, and timing.
Specifically, SubER computes the number of word edits and block/line edits required to match the reference, where hypothesis and reference words are allowed to match only within subtitle blocks that overlap in time.
This definition highlights how SubER is sensitive only to major errors in terms of timing, as it solely checks whether two blocks overlap, even if only for a few milliseconds. 
This limitation motivates our proposal of SubSONAR, a new metric designed to be more sensitive to time shifts and, consequently, more suitable for evaluating the quality of subtitle timestamps.

To this aim, we leverage the multimodal and multilingual SONAR model \citep{Duquenne:2023:sonar_arxiv}, designed 
to generate
\mn{sentence embeddings 
within a shared multimodal (text and audio)  semantic space for all languages.}
Specifically, we calculate the cosine similarity between the SONAR embeddings of the text within a subtitle block and its corresponding audio, determined by the timestamp of the block. Subsequently, we average the similarity scores across all subtitle blocks, which results in a single score in the $[-1, 1]$ range. Being SONAR trained to capture both text and audio semantics, higher SubSONAR scores indicate better alignment between text and audio content, which is influenced by both translation quality and timestamp accuracy.
Nevertheless, as revealed by the empirical validation presented in the following sections (\S\ref{sec:results},\ref{sec:manual-eval}), SubSONAR exhibits higher sensitivity to timing accuracy than to translation quality.

\section{Automatic Evaluation}
\label{sec:results}

In this section, we evaluate our solutions automatically through 
comparative
experiments
in different resource conditions and language settings. First, we validate the adoption of the joint CTC rescoring and compare the timestamp estimation methods introduced in \S\ref{sec:timestamp-est} using a multilingual system trained and tested on all the 7 language pairs (en$\rightarrow$\{de,es,fr,it,nl,pt,ro\}) of MuST-Cinema (\S\ref{subsec:res-ts-est-methods}). Then, we confirm their strength by 
\mn{comparing}
two bilingual systems (en-de and en-es) trained in the high resource conditions of 
\mn{the IWSLT 2023 subtitling track \citep{agrawal-etal-2023-findings}}
with the \mg{results reported for the} current state of the art \mg{in direct AS} and production tools on several test sets (\S\ref{subsec:tacl-compare}). Lastly, we demonstrate that our solutions close the gap with the cascade approach, 
outperforming the best 
\mn{IWSLT cascade models}
on the 4 publicly available validation sets (\S\ref{subsec:iwslt-compare}). 
Full experimental settings and details about test and training data
are 
given
in Appendix \ref{sec:exp-sett} to ensure the reproducibility of our work.

\subsection{Timestamp Estimation Methods}
\label{subsec:res-ts-est-methods}

\begin{table}[t]
\small
\setlength{\tabcolsep}{2.75pt}
    \centering
    \begin{tabular}{l|ccccccc|c}
        \textbf{decoding} & \textbf{de} & \textbf{es} & \textbf{fr} & \textbf{it} & \textbf{nl} & \textbf{pt} & \textbf{ro} & \textbf{AVG}  \\
        \hline
        standard & 20.9 & 34.3 & 27.9 & 28.6 & 30.9 & 35.1 & 30.1 & 29.7 \\
        joint CTC & \textbf{21.8} & \textbf{35.2} & \textbf{28.6} & \textbf{28.9} & \textbf{31.1} & \textbf{35.7} & \textbf{30.8} & \textbf{30.3} \\
    \end{tabular}
    \caption{Translation quality results measured by AS-BLEU ($\uparrow$) with and without joint CTC decoding for all the 7 languages of MuST-Cinema.}
    \label{tab:decoding-bleu}
\end{table}

\begin{table*}
\small
\setlength{\tabcolsep}{3pt}
    \centering
    \begin{tabular}{l|c:c||cc|cc|cc|c||cc|cc|cc|c}
        \multirow{3}{*}{\textbf{Model}} & \multirow{3}{*}{align.} & \multirow{3}{*}{joint} & \multicolumn{7}{c||}{\textbf{SubER} ($\downarrow$)} & \multicolumn{7}{c}{\textbf{SubSONAR} ($\uparrow$)} \\
        \cline{4-17}
        & & & \multicolumn{2}{c|}{MSTCIN} & \multicolumn{2}{c|}{ECSC} & \multicolumn{2}{c|}{EPI} & \multirow{2}{*}{AVG} & \multicolumn{2}{c|}{MSTCIN} & \multicolumn{2}{c|}{ECSC} & \multicolumn{2}{c|}{EPI} & \multirow{2}{*}{AVG} \\
        \cline{4-9} \cline{11-16}
        & & & de & es & de & es & de & es & & de & es & de & es & de & es & \\
        \hline
        \citet{10.1162/tacl_a_00607} & LEV & \no & 59.9 & 46.8 & 59.9 & 52.7 & 80.3 & 72.3 & 62.0 & \multirow{2}{*}{-} & \multirow{2}{*}{-} & \multirow{2}{*}{-} & \multirow{2}{*}{-} & \multirow{2}{*}{-} & \multirow{2}{*}{-} & \multirow{2}{*}{-}\\
        Best Production* & na & na & 61.5 & 51.3 & 59.0 & 49.7 & 78.1 & \textbf{68.6} & 61.4 & & & & & & \\
        \hline
        \multirow{3}{*}{This work} & \multirow{2}{*}{LEV} & \no & 58.0 & 48.6 & 59.6 & 49.9 & 78.5 & 70.2 & 60.8 & .656 & .707 & .691 & .713 & .670 & .697 & .689 \\
         &  & \yes & 56.5 & 45.1 & 58.9 & 49.0 & 77.6 & 69.7 & 59.5 & .668 & .712 & .693 & .722 & .676 & .701 & .695 \\
         \cdashline{2-17}
         & DTW & \yes & 56.3 & \textbf{44.7} & \textbf{58.5} & \textbf{48.9} & \textbf{77.3} & 69.7 & \textbf{59.2} & \textbf{.733} & .780 & \textbf{.742} & .778 & .733 & .763 & .755 \\
         & SBAAM & \yes & \textbf{56.2} & \textbf{44.7} & 58.6 & \textbf{48.9} & \textbf{77.3} & 69.6 & \textbf{59.2} & .728 & \textbf{.781} & .740 & \textbf{.785} & \textbf{.734} & \textbf{.770} & \textbf{.756} \\
    \end{tabular}
    \caption{SubER (cased) and SubSONAR results of our high-resource models with and without joint CTC generation (joint) and with the LEV, DTW or \textsc{SBAAM} timestamp alignment methods (align.), compared with previous work and production tools on MuST-Cinema (MSTCIN), EC Short Clips (ECSC) and EP Interviews (EPI) en-de and en-es. *~Best results of the production tools reported by \citet{10.1162/tacl_a_00607} for every language and test set.}
    \label{tab:prevworks}
\end{table*}

Table \ref{tab:mustcinema-suber-subsonar} 
\mn{shows}
the results of our multilingual system with the proposed timestamp estimation methods, comparing them to the Levenshtein-based approach (LEV) of \citet{10.1162/tacl_a_00607}. Notice that we employ the same model for all rows, meaning that the outputs vary solely in terms of block timings, except for the ablation row (- joint CTC) where we analyze the impact of the joint CTC rescoring on 
the subtitle
quality. 

First of all, we notice a notable SubER decrease with joint CTC rescoring (-0.9 on average), while the different timing strategies have a lower impact on it (improvements 
\mn{remain below}
0.4 on average), with the two attention-based methods (ATTN DTW and \textsc{SBAAM}) achieving the same scores. On the contrary, SubSONAR is not significantly affected by the decoding strategy but exhibits 
\mn{increasingly higher scores}
as block timings become more accurate. The substantial gains ($>$6\% relative improvement) of SubCTC over LEV underscore the importance of directly estimating timestamps on the subtitle blocks, thereby avoiding inherent error propagation of mapping them from the caption blocks (as done in LEV). Consistently with the SubER scores, the attention-based methods lead to superior scores, with \textsc{SBAAM} emerging as the best timestamp estimation strategy, outperforming ATTN DTW by a limited, yet consistent, margin across all languages. Overall, \textsc{SBAAM} improves SubSONAR by $~$11.8\% over the current state of the art (LEV) for direct subtitling. 

In addition to certifying the effectiveness of our solutions, these results demonstrate that SubER is more sensitive to translation quality than timestamp accuracy. Comparing the results in Table \ref{tab:mustcinema-suber-subsonar} with those in Table \ref{tab:decoding-bleu}, it is evident that the substantial gains achieved through joint CTC rescoring are proportional to the improvements in terms of AS-BLEU,\footnote{AS-BLEU, computed with SubER tool \citep{suber_toolkit}, is calculated with the popular sacreBLEU metric \citep{post-2018-call} after aligning blocks of the hypothesis and reference with the minimum Levenshtein distance method \citep{matusov-etal-2005-evaluating}.} which measures the pure translation quality. Notably, the languages exhibiting the smallest decrease in SubER coincide with those where BLEU scores are closer (it and nl), while the language with the highest BLEU increase (es) also demonstrates the widest margin in terms of SubER. Conversely, timestamp accuracy shows limited effects on SubER scores, despite the important difference between methods, as attested by SubSONAR scores and validated by our manual analysis (\S\ref{sec:manual-eval}).

Lastly, we verify the effect of the proposed solutions in terms of subtitle conformity, namely the percentage of blocks that respect character-per-line (CPL) and character-per-second (CPS) limits. Such limits, respectively set to 42 and 21 in TED guidelines,\footnote{\url{https://www.ted.com/participate/translate/subtitling-tips}} ensure that subtitles can be understood by users without excessive cognitive effort.
As it has been demonstrated that the joint CTC rescoring solves the end detection problem of purely attentional models and promotes hypotheses with correct length \citep{yan-etal-2023-ctc}, we hypothesized that its use could improve both CPL and CPS metrics. However, inconsistent outcomes across various settings and languages (shown in Appendix \ref{app:cpl_cps}) dispute this hypothesis, 
whereas the attention-based timing methods consistently improve the CPS, with \textsc{SBAAM} being superior also in this respect.

\begin{table*}
\small
    \centering
    \begin{tabular}{l||cc|cc|cc|cc||cc}
        \hline
        \multicolumn{11}{c}{\textbf{en-de}} \\
        \hline
        \multirow{3}{*}{\textbf{Model}} & \multicolumn{2}{c|}{\textit{TED}} & \multicolumn{2}{c|}{\textit{EPTV}} & \multicolumn{2}{c|}{\textit{ITV}} & \multicolumn{2}{c||}{\textit{PELOTON}} & \multicolumn{2}{c}{\textbf{\textit{AVG}}} \\
        \cline{2-11}
        & \multicolumn{2}{c|}{SubER} & \multicolumn{2}{c|}{SubER} & \multicolumn{2}{c|}{SubER} & \multicolumn{2}{c||}{SubER} & \multicolumn{2}{c}{SubER} \\
        & cased & uncased & cased & uncased & cased & uncased & cased & uncased & cased & uncased \\
        \hline
        Best Cascade & - & 63.0 & - & 78.7 & - & 83.6 & - & 87.6 & - & 78.2 \\
        Best Direct & 69.4 & - & 80.6 & - & 83.7 & - & 79.1 & - & 78.2 & - \\
        \hdashline
        This work & \textbf{61.6} & \textbf{62.1} & \textbf{78.7} & \textbf{78.3} & \textbf{80.0} & \textbf{80.7} & \textbf{75.6} & \textbf{78.2} & \textbf{74.0} & \textbf{74.8} \\
        \hline
        \multicolumn{11}{c}{\textbf{en-es}} \\
        \hline
        \multirow{3}{*}{\textbf{Model}} & \multicolumn{2}{c|}{\textit{TED}} & \multicolumn{2}{c|}{\textit{EPTV}} & \multicolumn{2}{c|}{\textit{ITV}} & \multicolumn{2}{c||}{\textit{PELOTON}} & \multicolumn{2}{c}{\textbf{\textit{AVG}}} \\
        \cline{2-11}
        & \multicolumn{2}{c|}{SubER} & \multicolumn{2}{c|}{SubER} & \multicolumn{2}{c|}{SubER} & \multicolumn{2}{c||}{SubER} & \multicolumn{2}{c}{SubER} \\
        & cased & uncased & cased & uncased & cased & uncased & cased & uncased & cased & uncased \\
        \hline
        Best Cascade & - & 48.8 & - & \textbf{70.2} & - & 82.1 & - & \textbf{79.0} & - & 70.0 \\
        Best Direct & 52.5 & - & 73.7 & - & 82.2 & - & 80.3 & - & 72.2 & - \\
        \hdashline
        This work & \textbf{49.5} & \textbf{47.5} & \textbf{73.1} & 71.0 & \textbf{79.1} & \textbf{79.5} & \textbf{79.3} & 80.8 & \textbf{70.2} & \textbf{69.7} \\
        \hline
    \end{tabular}
    \caption{SubER ($\downarrow$) comparison with the best cascade (AppTek) and direct (FBK) models trained on constrained conditions from the IWSLT 2023 Evaluation Campaign on automatic subtitling for en-de and en-es validation sets.
    }
    \label{tab:iwslt}
\end{table*}

\subsection{High-Resource Conditions}
\label{subsec:tacl-compare}

To further validate the robustness of our findings, we experiment in high-resource conditions with bilingual systems (en-de and en-es) and compare our solutions with the state of the art. Besides TED talks, we leverage two additional test sets, European Commission (EC) Short Clips and European Parliament (EP) Interviews \citep{10.1162/tacl_a_00607}, to cover different domains and data conditions: multi-speaker short informative videos about various topics with background music, and interviews with non-verbatim subtitles. The results of our solution are reported in Table \ref{tab:prevworks} and compared with the current best scores published on these test sets.

Upon comparing rows 1 and 3, we observe that our baseline model, even without joint CTC rescoring and replicating the LEV method, outperforms previous results across nearly all test sets, with the sole exception of MuST-Cinema en-es, affirming its competitiveness and reinforcing our findings. Furthermore, we observe consistent trends and similar improvements as in the previous section when applying joint CTC rescoring and employing \textsc{SBAAM} for timestamp estimation. On average across all languages and test sets, joint CTC rescoring decreases the SubER by 1.3 with negligible effects on SubSONAR, while \textsc{SBAAM} yields limited improvement (0.3) in SubER but a substantial relative increase of 8.8\% in SubSONAR.

Overall, our proposed solution (last row) emerges as the best one on all test sets, except for EP Interviews en-es, where the best production system reported by \citet{10.1162/tacl_a_00607}\footnote{Note that, as we did not test such production systems, they may have been improved at the time of writing this paper.} has a better SubER (-1.0). On average, our model is 2.2 SubER better than cherry-picking the best production system for each test set and language, and 1.6 SubER better than our baseline. To further confirm its strength, in the next section, we compare it with the best systems of the last IWSLT campaign.

\subsection{Is the Gap with Cascade Systems Closed?}
\label{subsec:iwslt-compare}

In this section, we compare our high-resource models with the best direct (FBK -- \citealt{papi-etal-2023-direct}) and the best cascade (AppTek -- \citealt{bahar-etal-2023-speech}) systems of the last IWSLT campaign \citep{agrawal-etal-2023-findings}, in the constrained data condition, meaning the models are trained on the same data as ours. As the references for the official test sets are not public, we present the results on the 4 validation sets released for the campaign in Table \ref{tab:iwslt}. For the sake of a fair comparison, we report the SubER with and without casing and punctuation, as \citet{bahar-etal-2023-speech} report the latter.

\mn{Even in this testing condition, our systems consistently outperform the others,} surpassing even the top-performing cascade model on both language pairs. 
The superiority of our solution is particularly pronounced in en-de, where it achieves the lowest SubER in all conditions by a large margin (-3.4 SubER on average over the best cascade and -4.2 over the direct). On en-es, our solution and the cascade are, instead, close, with our models emerging on TED and ITV, whereas the cascade models are better on on EPTV and PELOTON. On average, however, our solution results slightly superior in this language pair with a 0.3 SubER reduction.
All in all, our experiments evidence that our proposed solution can effectively close the gap between cascade and direct subtitling systems for the first time.

\section{Manual Evaluation}
\label{sec:manual-eval}
%
%
To corroborate the findings from our automatic evaluation, we conducted the first-ever manual evaluation of timestamp quality in subtitling. This evaluation was specifically conducted on the two language pairs addressed in our high-resource experiments (en-de, en-es).

To focus only
on timestamp quality, we compared different timestamp estimation techniques 
on the same automatic translations. Specifically, we selected the outputs of our high-resource systems with joint CTC rescoring and confronted the baseline LEV method and our proposed \textsc{SBAAM}. 

For each language pair, the manual evaluation set -- on which the two systems were run -- is composed of 
subsets of the EC Short Clips and MuST-Cinema test sets, 
for a total of 22 videos corresponding to approximately 1 hour of audio (see Appendix \ref{subsec:audio-selection} for details on the selected videos).

The evaluation was carried out by two annotators per language pair, who are proficient in English (C1) as well as native speakers or very proficient (C2) in the target language (German or Spanish).\footnote{The annotators were paid 22€/h gross, in accordance with the average salary of data annotators (\url{https://www.talent.com/salary?job=data+annotator}).}
We instructed the annotators with ad-hoc guidelines (described in Appendix \ref{subsec:he-guidelines}) to adjust the start, end, or both timestamps when the generated subtitles are not synchronized with the speech. To collect this information, we used ELAN \citep{wittenburg-etal-2006-elan}, an audio/video annotation tool commonly 
\mn{employed}
in the literature \citep{sloetjes-wittenburg-2008-annotation}.
To cope with the inherent assessors' subjectivity (i.e. 
\lb{being}
 more aggressive/tolerant 
\lb{in deciding} whether a timing is acceptable or has to be edited) the outputs of the two timestamp estimation methods were randomly assigned to the two annotators, ensuring that each one worked on all 22 audios
and annotated
half of the outputs from both methods. 
%
To prove the quality of the manual evaluation and better understand the difficulty of the task, for each language pair, 
25\%
of the test set was double-annotated. We calculated Cohen's Kappa \citep{cohen1960coefficient} to measure the inter-annotator
agreement on whether a start/end timestamp has to be edited or not.
%
The resulting values --
0.65 for en-de and 0.61 for en-es --
indicate a \enquote{substantial} agreement \citep{landis1977measurement}, 
confirming
both the feasibility of this new evaluation task and the reliability of its outcomes.

\begin{table}[t]
\small
    \centering
    \begin{tabular}{l|cccc}
       \hline
        \multicolumn{5}{c}{\textbf{en-de}} \\
       \hline
       \multirow{2}{*}{\textit{model}} & \multirow{2}{*}{\textit{time shift (ms)}} & \multicolumn{3}{c}{\textit{\% edited ts}} \\
       \cline{3-5}
        &  & start & end & avg \\
        \hline
       LEV & 544 $\pm$ 718 & 38.47 & 40.74 & 39.61 \\
       \texttt{SBAAM} & \textbf{321 $\pm$ 453} & \textbf{18.43} & \textbf{18.81} & \textbf{18.62} \\
       \hline
       \multicolumn{5}{c}{\textbf{en-es}} \\
       \hline
       \multirow{2}{*}{\textit{model}} & \multirow{2}{*}{\textit{time shift (ms)}} & \multicolumn{3}{c}{\textit{\% edited ts}} \\
       \cline{3-5}
        &  & start & end & avg \\
        \hline
       LEV & 520 $\pm$ 626 & 38.01 & 39.45 & 38.73 \\
       \texttt{SBAAM} & \textbf{347 $\pm$ 570} & \textbf{11.17} & \textbf{12.89} & \textbf{12.03} \\
       \hline 
    \end{tabular}
    \caption{Results of the manual evaluation in terms of time shift in milliseconds (mean and standard variation), and percentage of the number of edited timestamps (divided in start and end timestamp, and their average).}
    \label{tab:he}
\end{table}

Table \ref{tab:he} shows the results of the manual analysis in terms of start/end time shifts (in milliseconds) and percentage of modified timestamps.
The superiority of \textsc{SBAAM} over LEV is evident across all metrics, domains, and language pairs. The percentage of modified timestamps is less than half in en-de and 3 times lower in en-es. Given that correcting automatic timestamp errors is a major concern for professionals in their subtitling experience with AI-based systems \citep{koponen-etal-2020-mt-subtitling}, the reduced error rates of \textsc{SBAAM} 
\mn{are}
likely to significantly
alleviate post-editing 
\mn{effort.}
Furthermore, \textsc{SBAAM} exhibits a notably lower average time shift (33-40\%), indicating not only less frequent but also less severe errors. 
The lower average \mn{time} shift is complemented by a significantly lower standard deviation, which further evidences that \textsc{SBAAM} errors are less likely to be large.

\mg{Lastly, in Appendix \ref{subsec:filtered-he} we present the results of this manual analysis after excluding edits under 120 ms. This evaluation caters to user experience, as users typically perceive audio-visual stimuli under 120 ms as instantaneous \citep{EFRON197057}, while the annotators -- aided by ELAN that shows the text and waveform -- made numerous fine-grained timestamp edits (even $<$20 ms).}
This 
evaluation accentuates the differences between the two methods and shows that SBAAM requires editing for only 12\% of timestamps in en-de and 9\% in en-es.   


We can conclude that our manual analysis 
\lb{not only confirms}
the results obtained through the automatic SubSONAR metric,
but also 
\lb{further highlights}
the efficacy of our proposed timestamp estimation method and its substantial superiority over previous solutions.




\section{Conclusions}


\mn{In recent years, research on automatic subtitling has shifted toward direct models that do not rely on the transcription of the input audio. The potential advantages of a transcription-free approach have motivated the use of direct models for the translation and segmentation steps, leaving unexplored the direct generation of timestamps on the target  
text.
In this paper, we filled this gap by introducing the first direct 
system that does not require the generation of transcripts/captions in any phase of the process, including subtitle timestamp estimation. With the introduction of a new metric, SubSONAR, dedicated to evaluating timestamp quality, and experiments on different domains and 7 language pairs, we demonstrated the effectiveness of our solution, which was further validated by a dedicated manual evaluation. Lastly, we showed that our model closed the gap with the cascade paradigm, approaching and even outperforming the best cascade architectures from the last IWSLT campaign, thereby setting a new state of the art for most of the publicly available benchmarks.}

\section*{Limitations}
While our direct AS model achieves significant advancements, there are still some limitations or problems that have not been addressed in this work and should be the focus of further research on the topic.

First of all, although our models are easily applicable to unwritten languages (the only required change is either to use speech units as targets for the CTC compression module or to remove the module itself), our experiments have not included unwritten languages due to the lack of available benchmarks. Moreover, for the same reason, we have not experimented with source languages different from English. Despite this, we do not foresee any specific problem that could arise in this condition, except for a slight drop in translation quality if the CTC compression and its auxiliary loss are removed \citep{pmlr-v162-zhang22i}.

Another limitation of this work regards the model compliance with constraints posed on CPL and CPS. While our model is trained on subtitles that mostly adhere to spatio-temporal requirements, no specific strategies are adopted for their actual fulfillment. Modifying the model or the training strategy to consistently meet these constraints, especially in complex audiovisual content, should be the topic of future works.

Regarding SubSONAR, its language coverage is currently limited to the languages supported by the SONAR speech encoder.

Lastly, due to the large computational costs required and for the sake of a fair comparison with previous works, we did not experiment with using our proposed method on top of large foundational ST models such as SeamlessM4T \citep{seamless2023} and Whisper \citep{whisper}, although the latter can not be tested in our settings as it does not support translating from English into other languages. To adopt these models for the subtitling task, line and block boundary tokens should be added to their vocabulary and these models should be then fine-tuned on subtitling data (i.e. translations with line and block boundaries). No other modification is needed as \textsc{SBAAM} requires only looking at the cross-attention. Optionally, a CTC on Target module has to be trained on top of their encoder for the joint CTC rescoring. This line of research represents a natural next step of this work.



\section*{Acknowledgements}
We acknowledge the support  of the PNRR project FAIR -  Future AI Research (PE00000013),  under the NRRP MUR program funded by the NextGenerationEU.
The work is co-funded by the European Union under the project {\it AI4Culture: An AI platform for the cultural heritage data space} (Action number 101100683).
We acknowledge the CINECA award under the ISCRA initiative, for the availability of high-performance computing resources and support.

\bibliography{custom}

\clearpage

\appendix

\section{Experimental Settings}
\label{sec:exp-sett}

\subsection{Data}
For the multilingual model, we leverage all the 7 language \mc{pairs} of MuST-Cinema v1.1 \citep{karakanta-etal-2020-must}, namely: English to German, Spanish, French, Italian, Dutch, Portuguese, and Romanian.
These texts already contain line and block segmentation, i.e., \texttt{<eol>} and \texttt{<eob>} tags are present in both transcripts and translations. 

For the bilingual models, we use the same datasets of \citep{papi-etal-2023-direct}, encompassing most of the training data admitted by the IWSLT 2023 Evaluation Campaign on automatic subtitling. We collect all the available ST corpora, namely MuST-Cinema, EuroParl-ST \citep{europarlst}, and CoVoST v2 \citep{wang2020covost}. Also, we leverage most of the available ASR datasets (CommonVoice \citep{ardila-etal-2020-common}, LibriSpeech \citep{librispeech}, TEDLIUM v3 \citep{tedlium}, and VoxPopuli \citep{wang-etal-2021-voxpopuli}), by automatically translating the transcripts into the two target languages (German and Spanish) using the NeMo 
MT models.\footnote{Publicly available at: \url{https://docs.nvidia.com/deeplearning/nemo/user-guide/docs/en/main/nlp/machine_translation/machine_translation.html}}

\texttt{<eol>} and \texttt{<eob>} tags are added to both transcripts and translations of all datasets, except for MuST-Cinema that already include them, using the multimodal segmenter by \citet{papi-etal-2022-dodging}.

\subsection{Model and Training}
Our systems are implemented with fairseq-ST \cite{wang2020fairseqs2t} using the default settings unless specified otherwise. The input comprises 80 Mel-filterbank audio features extracted every 10 milliseconds, employing a sample window of 25. The input features are then preprocessed with two 1D convolutional layers with stride 2, reducing input length by a factor of 4. 

\begin{table}[!ht]
    \centering
    \small
    \begin{tabular}{c|c}
    \hline
        \multicolumn{2}{c}{\textbf{Encoder}} \\
        \hline
        Layer type & Conformer \\
        Total number of layers & 12 \\ 
        $N_a$ layers & 8 \\
        $N_s$ layers & 4 \\
        Embedding dimension & 512 \\
        FFN dimension & 2,048 \\
        Convolutional Module kernel & \multirow{2}{*}{31} \\
        size (point- and depthwise) & \\
    \hline
        \multicolumn{2}{c}{\textbf{Decoder}} \\
        \hline
        Layer type & Transformer \\
        $N_d$ layers & 6 \\ 
        Embedding dimension & 512 \\
        FFN dimension & 2,048 \\
    \hline
    \end{tabular}
    \caption{Hyperparameters for the proposed model.}
    \label{tab:model-settings}
\end{table}

The model architecture follows an encoder-decoder design, consisting of a Conformer encoder \citep{gulati20_interspeech} and a Transformer decoder \citep{transformer}, with a total number of 133M for both multilingual and bilingual models.
The vocabularies are based on unigram SentencePiece \citep{kudo-2018-subword}, with size 8,000 for the English source and 16,000 for the target (either German, Spanish, or multilingual). 
Specific hyperparameters are presented in Table \ref{tab:model-settings}.

\begin{table}[!ht]
    \small
    \centering
    \begin{tabular}{c|c}
    \hline
        Optimizer & AdamW \\
        Optimizer Momentum & $\beta_1,\beta_2=0.9,0.98$ \\ 
        Source CTC weight ($\lambda_1$) & 1.0 \\
        Target CTC weight ($\lambda_2$) & 2.0 \\
        CE weight ($\lambda_3$) & 5.0 \\
        CE label smoothing & 0.1 \\
        Learning Rate scheduler & Noam \\
        Learning Rate & 2e-3 \\
        Warmup steps & 25,000 \\
        Weight Decay & 0.001 \\
        Dropout & 0.1 \\
        Clip Normalization & 10.0 \\
        Training steps & 200,000 \\
        Maximum tokens & 40,000 \\
        Update frequency & 2 \\
        \hline
    \end{tabular}
    \caption{Model detailed training settings. The total batch size (in number of tokens) is obtained by multiplying maximum tokens for update frequency.}
    \label{tab:train-settings}
\end{table}

Label-smoothed CE is computed after the decoder, target CTC loss is computed on the output of the semantic encoder (full encoder), and the source CTC loss is computed on the output of the acoustic encoder. 
The values of the weights for all three losses ($\lambda_1,\lambda_2,\lambda_3$) are set to (1.0, 2.0, 5.0) according to \citep{yan-etal-2023-ctc}.
Utterance-level Cepstral Mean and Variance Normalization (CMVN) and SpecAugment \cite{Park2019} are applied during training and segments longer than 30 seconds are filtered out (fairseq-ST default) to avoid excessive VRAM requirements. All model checkpoints are obtained by averaging the last 7 checkpoints obtained from the training. All trainings are executed on 4 NVIDIA Ampere GPU A100 (64GB VRAM). 

For the multilingual case, we train the model in one step on all the MuST-Cinema languages by pre-pending the language tag to the subtitle texts, as already explained in \S\ref{sec:model}. For the bilingual case, we first train the model on the ST and the machine-translated ASR datasets without \texttt{<eol>} and \texttt{<eob>} tags, and then we continue the training starting from the averaged checkpoint by including \texttt{<eol>} and \texttt{<eob>} tags in the texts. For both cases (and steps, in the case of bilingual models), we use the training settings provided in Table \ref{tab:train-settings}.

For inference, we set the beam size to 5 and, according to \citep{yan-etal-2023-ctc}, the joint CTC decoding weight $\alpha$ to 0.2. For the SBAAM timestamp estimation method, we extract the cross-attention from the 4\textsuperscript{th} layer and average the scores across the attention heads, following \citep{papi-etal-2023-attention}.

\section{CPL and CPS Conformity}
\label{app:cpl_cps}

\begin{table}[t]
\small
\setlength{\tabcolsep}{2.75pt}
    \centering
    \begin{tabular}{l|ccccccc|c}
         \textbf{decoding} & \textbf{de} & \textbf{es} & \textbf{fr} & \textbf{it} & \textbf{nl} & \textbf{pt} & \textbf{ro} & \textbf{AVG} \\
         \hline
         standard & 89.6 & \textbf{94.7} & \textbf{91.7} & 88.7 & 84.1 & 89.1 & 92.0 & 90.0 \\
        joint CTC & \textbf{90.1} & 94.6 & 91.0 & \textbf{89.3} & \textbf{85.1} & \textbf{89.4} & \textbf{93.7} & \textbf{90.5} \\
        \hline
    \end{tabular}
    \caption{Results of the CPL conformity ($\uparrow$) in percentage (\%) with and without joint CTC decoding for all the 7 languages of MuST-Cinema.}
    \label{tab:cpl-multi}
\end{table}

\begin{table}[t]
\small
\setlength{\tabcolsep}{4pt}
    \centering
    \begin{tabular}{c||cc|cc|cc|c}
         \multirow{2}{*}{\textbf{decoding}} & \multicolumn{2}{c|}{\textbf{MSTCIN}} & \multicolumn{2}{c|}{\textbf{ECSC}} & \multicolumn{2}{c|}{\textbf{EPI}} & \multirow{2}{*}{\textbf{AVG}} \\
        \cline{2-7}
         & de & es & de & es & de & es & \\
        \hline
         standard & 88.2 & \textbf{95.3} & \textbf{85.0} & 90.9 & \textbf{82.3} & \textbf{90.8} & \textbf{88.8} \\
         joint CTC & \textbf{88.3} & 94.6 & 84.0 & \textbf{91.1} & 82.2 & 89.7 & 88.3 \\
    \end{tabular}
    \caption{CPL conformity results ($\uparrow$) in percentage (\%) of our high-resource models with and without joint CTC generation on MuST-Cinema (MSTCIN), EC Short Clips (ECSC) and EP Interviews (EPI) en-de and en-es.}
    \label{tab:cpl-bi}
\end{table}

Table \ref{tab:cpl-multi} and \ref{tab:cpl-bi} report the CPL conformity of, respectively,  our multilingual and bilingual models with and without the joint CTC decoding strategy. As we can see, the results are very close and no clear and coherent trend across the different languages and test sets emerges. We can conclude that the decoding strategy does not significantly impact CPL conformity.

Switching to analyze the CPS conformity, Table \ref{tab:cps-multi} and \ref{tab:cps-bi} show the related percentages for the multilingual and bilingual models. In this case, we do not only study the decoding strategy but also the timing estimation method used, as CPS is influenced both by the generated subtitles and the assigned timestamps. While also in this case the difference between the decoding strategies is limited, on average the joint CTC rescoring leads to a lower conformity in both cases. Looking at the differences between the timing estimation methods (Table \ref{tab:cps-multi}), we notice that SubCTC is by far the worst method (with a $\sim$7\% degradation). ATTN DTW and \textsc{SBAAM}, instead, lead to subtitles with higher conformity than the LEV baseline, although \textsc{SBAAM} is consistently slightly superior in all languages. The benefits of \textsc{SBAAM} over LEV are confirmed also by the bilingual systems, where \textsc{SBAAM} has a 0.5\% higher compliance on average, although the gains are not coherent across the two language pairs.

We can conclude that we can reject the hypothesis about the potential benefits of the joint CTC rescoring in producing outputs of the correct length. The better timings assigned by the timestamp estimation methods, instead, provide little benefits in terms of CPS conformity.

\begin{table}[t]
\small
\setlength{\tabcolsep}{2.2pt}
    \centering
    \begin{tabular}{l|ccccccc|c}
        \textbf{model} & \textbf{de} & \textbf{es} & \textbf{fr} & \textbf{it} & \textbf{nl} & \textbf{pt} & \textbf{ro} & \textbf{AVG} \\
         \hline
        LEV & 74.1 & 77.4 & 68.7 & 76.9 & 79.2 & 79.8 & \textbf{84.1} & 77.2 \\
        \quad- joint CTC & 73.6 & 75.8 & 67.5 & 77.8 & 77.5 & 79.8 & 83.9 & 76.6 \\
        \hdashline
        SubCTC & 68.3 & 70.6 & 63.7 & 69.6 & 72.9 & 72.8 & 77.1 & 70.7 \\
        ATTN DTW & 75.6 & 79.3 & 71.3 & 78.0 & 80.5 & 81.4 & 83.8 & 78.6 \\
        SBAAM & \textbf{75.7} & \textbf{79.7} & \textbf{72.5} & \textbf{78.5} & \textbf{81.7} & \textbf{82.1} & 84.0 & \textbf{79.2} \\
        \hline
    \end{tabular}
    \caption{Results of the timestamp estimation methods in terms of CPS conformity ($\uparrow$) in percentage (\%) for all the 7 languages of MuST-Cinema.}
    \label{tab:cps-multi}
\end{table}

\begin{table}[t]
\small
\setlength{\tabcolsep}{2.5pt}
    \centering
    \begin{tabular}{c:c||cc|cc|cc|c}
         \multirow{2}{*}{\textbf{align.}} & \multirow{2}{*}{\textbf{joint}} & \multicolumn{2}{c|}{\textbf{MSTCIN}} & \multicolumn{2}{c|}{\textbf{ECSC}} & \multicolumn{2}{c|}{\textbf{EPI}} & \multirow{2}{*}{\textbf{AVG}} \\
        \cline{3-8}
         & & de & es & de & es & de & es & \\
        \hline
         \multirow{2}{*}{LEV} & \no & \textbf{78.8} & \textbf{81.7} & 82.3 & 83.7 & 74.1 & 77.4 & \textbf{79.7} \\
          & \yes & 77.8 & 76.0 & \textbf{82.8} & 85.2 & \textbf{74.6} & 76.6 & 78.8 \\
         \cdashline{2-9}
         \textsc{SBAAM} & \yes & 76.7 & 77.5 & 82.0 & \textbf{87.4} & 73.4 & \textbf{78.7} & 79.3 \\
    \end{tabular}
    \caption{CPS conformity results ($\uparrow$) in percentage (\%) of our high-resource models with and without joint CTC generation (joint) and with the LEV or \textsc{SBAAM} timestamp alignment methods (align.) on MuST-Cinema (MSTCIN), EC Short Clips (ECSC) and EP Interviews (EPI) en-de and en-es. }
    \label{tab:cps-bi}
\end{table}

\section{Human Evaluation}

\subsection{Audio Selection}
\label{subsec:audio-selection}

Table \ref{tab:audio-selection} lists the audios included in the manual evaluation carried out by the annotators.

\begin{table}[!ht]
\small
    \centering
    \begin{tabular}{cc|c}
        \textbf{Test Set} & \textbf{Audio Name} & \textbf{Duration} \\
        \hline
         \multirow{16}{*}{ECSC} & I118982 & 0:02:38 \\
            & I200637 & 0:01:30 \\
            & I203453 & 0:01:38 \\
            & I205895 & 0:01:31 \\
            & I207338 & 0:01:29 \\ 
            & I207340 & 0:01:35 \\
            & I207805 & 0:01:52 \\
            & I212173* & 0:03:01 \\
            & I207806 & 0:02:00 \\
            & I207807 & 0:01:54 \\
            & I207810 & 0:01:59 \\
            & I207813 & 0:02:21 \\
            & I207814 & 0:01:58 \\
            & I209528 & 0:02:08 \\
            & I211037 & 0:01:58 \\
            & I212173* & 0:03:01 \\
            \hdashline
            \multirow{6}{*}{MSTCIN} & 13518 & 0:05:23 \\
                & 20008 & 0:05:03 \\
                & 28521* & 0:04:41 \\
                & 22979 & 0:05:02 \\
                & 23129 & 0:05:35 \\
                & 28521* & 0:04:41 \\
            \hline
            & & 1:02:58 \\
    \end{tabular}
    \caption{Audio selection for the manual analysis from EC Short Clips (ECSC) and MuST-Cinema (MSTCIN). * Repeated two times for computing the inter-annotator agreement for the two methods.}
    \label{tab:audio-selection}
\end{table}


\subsection{Guidelines for Subtitle Timestamps Evaluation}
\label{subsec:he-guidelines}

The following italic text has been given to the annotators as guidelines for their work.

\textit{You are asked to check if the source language speech and its corresponding translated subtitles are synchronized, that is if the subtitle remains on screen for the right amount of time with respect to the corresponding speech.}

\textit{If the timing is wrong, move the timestamp of each subtitle such that the timing of the subtitle matches the timing of the audio. The timestamp can be adjusted at the beginning by changing the start timestamp, at the end by changing the end timestamp, or both. Change the timestamp only when the content is not aligned or only partially aligned. Please notice that the content of the subtitles, even if not correct, must not be changed. Only modifications to the timestamps are allowed.}

\textit{To summarize, please follow these guidelines:}
\begin{itemize}
    \item \textit{Adjust the timestamp such that the subtitle is synchronized with the corresponding audio: for this purpose, you can adjust the timestamp either at the beginning, at the end, or both;}
    \item \textit{Make only necessary changes to align the partially or completely misaligned subtitle with the audio;}
    \item \textit{Make only necessary changes to align the partially or completely misaligned subtitle with the audio;}
    \item \textit{Do not change the subtitle content.}
\end{itemize}

\textit{You will be given 22 videos to annotate and the work consists of around 6 hours of annotation. Please annotate the audio following the given order (from 1 to 22). To carry out your evaluation work at best, you should work on around 5 minutes of audio without interruptions. Then, you should take a break of around 5 minutes. We suggest to take a break after file numbers: 03, 06, 08, 11, 14, 16, 17, 18, 19, 20, 21, 22.}

\textit{It can happen that you have to annotate an audio that you have already annotated, please annotate it as if it were the first time.}

\begin{figure*}[!ht]
\centering
    \includegraphics[width=0.9\textwidth]{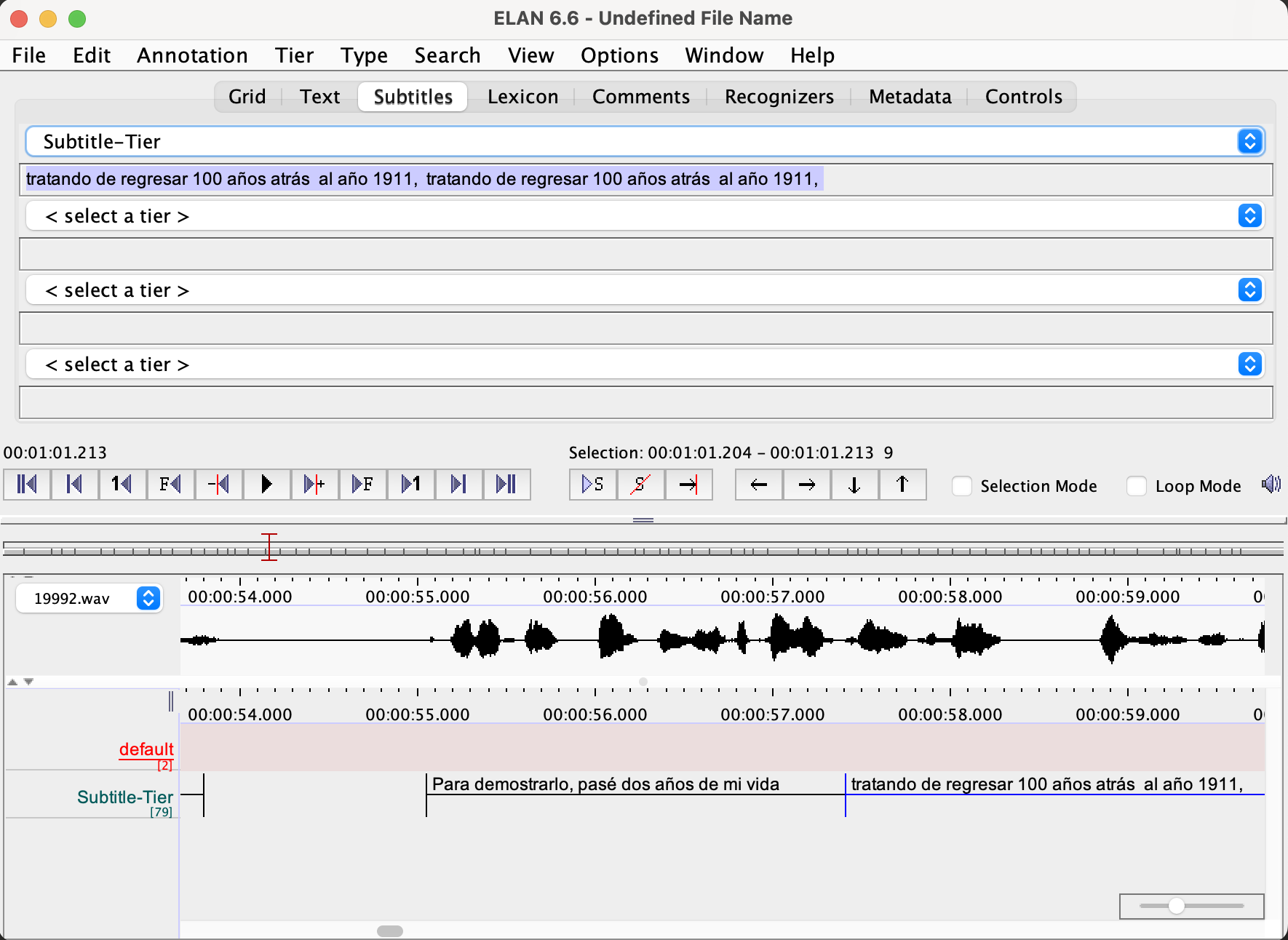}
    \caption{ELAN interface.}
    \label{fig:elan}
\end{figure*}

In addition, specific guidelines on the usage of the annotation tool, ELAN, were provided (Figure \ref{fig:elan} shows the interface). An initial training day, conducted with all the annotators and consisting of 3 hours of work, was held to explain the guidelines to the annotators, annotate a pilot audio, and discuss the results and common questions. This, with the 6 hours of annotation, resulted in 9 hours of work.

\subsection{Results with perception filtering}
\label{subsec:filtered-he}

\begin{table}[t]
\small
    \centering
    \begin{tabular}{l|cccc}
       \hline
        \multicolumn{5}{c}{\textbf{en-de}} \\
       \hline 
       \multirow{2}{*}{\textit{model}} & \multirow{2}{*}{\textit{time shift}} & \multicolumn{3}{c}{\textit{\% edited ts}} \\
       \cline{3-5}
        &  & start & end & avg \\
        \hline
       LEV & 606 $\pm$ 743 & 33.18 & 36.58 & 34.88 \\
       \textsc{SBAAM} & \textbf{422 $\pm$ 503} & \textbf{12.95} & \textbf{13.52} & \textbf{13.24} \\
       \hline
       \multicolumn{5}{c}{\textbf{en-es}} \\
       \hline
       \multirow{2}{*}{\textit{model}} & \multirow{2}{*}{\textit{time shift}} & \multicolumn{3}{c}{\textit{\% edited ts}} \\
       \cline{3-5}
        &  & start & end & avg \\
        \hline
       LEV & 629 $\pm$ 1015 & 28.56 & 33.43 & 31.00 \\
       \textsc{SBAAM} & \textbf{460 $\pm$ 632} & \textbf{8.31} & \textbf{9.65} & \textbf{8.98} \\
       \hline 
    \end{tabular}
    \caption{Results of the manual evaluation in terms of time shift in milliseconds (mean and standard variation), and percentage of the number of modified timestamp (divided by start and end timestamp, and their average) with filtered shifts under 120 milliseconds.}
    \label{tab:filtered-he}
\end{table}

As works in the field of human perception \citep{EFRON197057,Venezia_Thurman_Matchin_George_Hickok_2016} have demonstrated that humans perceive acoustic and visual stimuli with duration inferior to 120ms as instantaneous, we re-compute the statistics of the manual analysis by filtering out time shifts of less than 120 ms.
With this change, the resulting inter-annotator agreement increases to a Cohen's Kappa of 0.71 for en-de, and 0.68 for en-es, which is again \enquote{substatial} in both languages but higher than the one obtained in \S\ref{sec:manual-eval}.

The results of the two timestamp estimation methods (LEV and \textsc{SBAAM}) with the filtered shifts are shown in Table \ref{tab:filtered-he}. First of all, we notice that the time shift mean of \textsc{SBAAM} is $\frac{2}{3}$ of LEV, with important differences also in terms of the standard deviation, resulting in a reduction of 240ms for en-de and 383ms for en-es. Therefore, with filtered shifts, the gap between the two methods is more exacerbated than that shown in Table \ref{tab:he}, although with similar conclusions. Regarding the percentage of edited timestamps, we observe a relevant decrease in the number of edits for both methods, with \textsc{SBAAM} achieving less than 9\% of edited timestamps in en-es. Also in this case, the superiority of the \textsc{SBAAM} method over LEV is widened, as the number of edits for LEV is slightly less than 3$\times$  those for \textsc{SBAAM} in en-de, and nearly 3.5$\times$ in en-es.



\end{document}